\documentclass[runningheads]{llncs}
\usepackage{graphicx}
\usepackage{graphicx}
\usepackage{amsmath}
\usepackage{amsfonts}
\usepackage{multirow}
\usepackage{algorithm}
\usepackage[noend]{algpseudocode}
\usepackage[caption=false]{subfig}
\newcommand{\donotdisplay}[1]{}

\graphicspath{{Figures/}}
\newtheorem{prop}{Proposition}

\begin{document}
\title{Powered Dirichlet Process for Controlling the Importance of ``Rich-Get-Richer'' Prior Assumptions in Bayesian Clustering}
\titlerunning{Powered Dirichlet Process}
%
%
\author{Gael Poux-Medard\inst{1}\orcidID{0000-0002-0103-8778} \and
Julien Velcin\inst{1}\orcidID{0000-0002-2262-045X} \and
Sabine Loudcher\inst{1}\orcidID{0000-0002-0494-0169}}
%
%
\institute{$^1$ Université de Lyon, ERIC EA 3083, France\\
\email{gael.poux-medard@univ-lyon2.fr}\\
\email{julien.velcin@univ-lyon2.fr}\\
\email{sabine.loudcher@univ-lyon2.fr}\\
}
\maketitle              

\begin{abstract}
One of the most used priors in Bayesian clustering is the Dirichlet prior. It can be expressed as a Chinese Restaurant Process. This process allows nonparametric estimation of the number of clusters when partitioning datasets. Its key feature is the ``rich-get-richer'' property, which assumes a cluster has an \textit{a priori} probability to get chosen linearly dependent on population. In this paper, we show that such prior is not always the best choice to model data. We derive the Powered Chinese Restaurant process from a modified version of the Dirichlet-Multinomial distribution to answer this problem. We then develop some of its fundamental properties (expected number of clusters, convergence). Unlike state-of-the-art efforts in this direction, this new formulation allows for direct control of the importance of the ``rich-get-richer'' prior.
\end{abstract}


\section{Introduction}
The notion of clustering has been initially introduced by anthropologists Driver and Kroeber in 1932 \cite{Driver1932} in the classification of human psychological traits. It has been later used successfully in a broad range of applications, ranging from scientific research to data compression, marketing, and medicine. Over the past decades, it also became a central problem in machine learning\footnote{As an illustration, scraping Google Scholar shows that the yearly number of publications containing the keyword ``Clustering'' averages to 250.000.}.

The Bayesian clustering approach received broad attention in the last years. A non-exhaustive list of application includes medicine, \cite{SEESdrugdrugInter}, natural language processing \cite{LDA2003,Yin2014}, genetics \cite{mcdowell2018,Jensen2008,Qin2003}, recommender systems \cite{AiroldiMMSBM,AntoniaScalableRS,IMMSBM}, sociology \cite{HumanPrefSBM,SergioSocialDilemma}, etc. The key idea is to simulate a corpus of independent observations by drawing them from a set of latent variables (clusters). Those clusters are each associated with a probability distribution on the observations, whose parameters are drawn from a prior distribution, as we will formulate mathematically later. Now, an often desirable property of Bayesian models is to make them nonparametric. In our case, it means that both the number of clusters and their associated distributions are inferred. A very popular prior that allows this is the Dirichlet process. It incorporates a chance for a new cluster to be created in the prior probability of a distribution  (often when an observation is not likely to be explained by existing clusters). Otherwise, the observation is associated with an existing cluster with a probability proportional to that cluster's population.

However, the Dirichlet process (and the related Pitman-Yor process) comes with a hypothesis on the way observations are allocated to various clusters: the \textit{rich-get-richer} property \cite{Ferguson1973}. As stated before, a new observation belongs to a cluster with a probability proportional to the number of observations already present in the cluster; large clusters have a greater chance to get associated with new observations. While this can be a relevant property in some cases, it implies a strong assumption on the way data is generated. It has already been pointed out \cite{welling2006} that there is a need for more flexible priors. 
Imagine sampling topics over time on Twitter: there is no specific reason for a new word to belong to a topic with a probability depending linearly on the topic size as in the regular Dirichlet and Pitman-Yor processes. For instance, due to temporal variations of topic usage: a small news outbreak might go unnoticed using a DP. Since the emerging cluster has a low population, the ``rich-get-richer'' assumption could lead to ignore the outbreak if it is not different enough from another existing topic. Following the same idea in spatial clustering, tiny clusters (at the level of cities, for instance) might go unnoticed at larger scales (world map, for instance) if the ``rich-get-richer'' assumption has too much influence (see Fig.\ref{fig-Antoni}). In these cases, one needs to weaken the ``rich-get-richer'' assumption to control clustering's level of detail --- to make it flatter.

Little effort has been put into exploring alternative forms of priors for nonparameteric Bayesian modeling. In the present work, we offer to address this problem by deriving a more general form of the Dirichlet process that explicitly controls the importance of the ``rich-get-richer'' assumption. Explicitly, we derive the Powered Chinese Restaurant Process (PCRP) that allows control of the ``rich-get-richer'' property while generalizing state-of-the-art works. We show that controlling the ``rich-get-richer'' prior of simple models yields better results on synthetic and real-world datasets.

\section{Background}
\label{SotA}
\subsection{Motivation}
\label{motivation}
This work is motivated by the need to control the ``rich-get-richer'' assumption's importance in Dirichlet process (DP) priors.
The ``rich-get-richer'' property of the DP may not always be the most suitable prior for modeling a given dataset. The usual motivation for using a DP prior is that a new observation has a probability of being assigned to any cluster proportional to its population in the absence of external information (such as inter-points distance in case of spatial clustering, for instance). However, this assumption might be wrong (see Introduction). External information is often more relevant to the clustering than the clusters' population; we then want the DP prior to be flatter so that the model relies less on it. 

Most state-of-the-art works rely on tuning a parameter $\alpha$ (see Eq.\ref{eq-CRP}) to get the ``right'' number of clusters (this parameter shifts the distribution of the number of clusters as $\mathbb{E}(K \vert N) \propto \alpha \log N$ with $K$ the number of clusters and $N$ the number of observations). However, we argue this is a bad practice in some cases. Imagine sampling topics over time on Twitter: there is no specific reason for topics to appear at a rate $\alpha \log N$ as in the regular DP. If we later consider more observations on the same dataset, the parameter $\alpha$ would need to be tuned again to correct the likely wrong evolution of $K$ as $\log N$. It makes little sense since the dataset is still about the same data type.
Moreover, there is no specific reason for a new word to belong to a topic with a probability depending linearly on the topic size as in the regular Dirichlet and Pitman-Yor processes. A prior that is too peaky on crowded clusters might then lead to irrelevant results (due to temporal variations of topics, for instance, see Introduction and Fig.\ref{fig-Antoni}). To alleviate those assumptions, we develop a more general form of the DP process allowing a natural control of the ``rich-get-richer'' property.

\subsection{Previous works}
\subsubsection{Dirichlet process}
A well-known metaphor for the Dirichlet process is referred to as ``Chinese restaurant''. The corresponding process is named ``Chinese Restaurant Process'' (CRP). It can be illustrated as follows: if a $n^{th}$ client arrives in a Chinese restaurant, she will sit at one of the $K$ already occupied table with a probability proportional to the number of persons already sat at this table. She can also go to a new table in the restaurant and be the first client to sit there with a probability inversely proportional to the total number of clients already sat at other tables.
It can be written formally as:
\begin{equation}
\label{eq-CRP}
    CRP (C_i = c \vert C_1, C_2, ..., C_{i-1}) = 
    \begin{cases}
    \frac{N_c}{\alpha + N} \text{ if c = 1, 2, ..., K}\\
    \frac{\alpha}{\alpha + N} \text{ if c = K+1}
    \end{cases}
\end{equation}
Where $c$ is the cluster chosen by the $i^{th}$ customer, $N_k$ is the population of cluster $k$, $K$ is the number of already occupied tables and $\alpha$ the concentration parameter.
When the number of clients goes to infinity, this process is equivalent to a draw from a Dirichlet distribution over an infinite number of clusters with an identical initial probability to get chosen proportional to $\alpha$. The form of Eq.\ref{eq-CRP} is helpful to understand the underlying dynamics of the process and the contribution of seminal works we will detail now. It can be shown that the expected number of clusters after $N$ observations evolves as $\log N$ \cite{Arratia1992}.

The two best-known variations of the regular Dirichlet process that address the ``rich-get-richer'' property control are the seminal Pitman-Yor process and the Uniform process. Each of them can be expressed in a similar form as Eq.\ref{eq-CRP}.

\subsubsection{Pitman-Yor process}
Following the Chinese Restaurant process metaphor, the Pitman-Yor process \cite{PitmanYor1997,Ishwaran2003} proposed to incorporate a \textit{discount} when a client opens a new table. 
Mathematically, the process can be formulated as:
\begin{equation}
\label{eq-PY}
    CRP (C_i = c \vert C_1, C_2, ..., C_{i-1}) = 
    \begin{cases}
    \frac{N_c - \beta}{\alpha + N} \text{ if c = 1, 2, ..., K}\\
    \frac{\alpha + \beta K}{\alpha + N} \text{ if c = K+1}
    \end{cases}
\end{equation}

The introduction of the parameter $\beta>0$ increases the probability of creating new clusters. A table with a low number of customers has significantly less chances to gain new ones, while the probability of opening a new table increases significantly. It can be shown that the number of tables evolves with the number of clients $N$ as $N^{\beta}$ \cite{Sudderth2009,Goldwater2011}. However, this process does not control the arguable ``rich-get-richer'' hypothesis \cite{welling2006}, since the relation to the population of a table remains linear; it only shifts this dependence of a value $\beta$. It makes so by creating clusters based on the number of existing clusters and the total number of observations, but not according to the population of already existing clusters. Those play the same role in the Pitman-Yor process as in the DP. The Pitman-Yor process thus comes with two limitations. First, since $\beta>0$, it cannot modify the process to generate fewer clusters. Second, the discount parameter does not modify the linear dependence on previous observations for cluster allocations --- rich still get richer; the prior is as peaky on large clusters as before. The present work offers to address those two limitations.

\subsubsection{Uniform process}
Another process that aims at breaking the ``rich-get-richer'' property is the Uniform process. It has been used in some occasions \cite{Jensen2008,Qin2003} without proper definition. More recently, it has been formalized and studied in comparison with the regular Dirichlet and Pitman-Yor processes \cite{wallach2010}. It can be written as follows:
\begin{equation}
\label{eq-UP}
    CRP (C_i = c \vert C_1, C_2, ..., C_{i-1}) = 
    \begin{cases}
    \frac{1}{\alpha + K} \text{ if c = 1, 2, ..., K}\\
    \frac{\alpha}{\alpha + K} \text{ if c = K+1}
    \end{cases}
\end{equation}
This formulation completely gets rid of the ``rich-get-richer'' property. The probability of a new client joining an occupied table is a uniform distribution over the number of occupied tables; it does not depend on the tables' population. In \cite{wallach2010}, it has been shown that the expected number of tables evolves with $N$ as $\sqrt{N}$. Removing the ``rich-get-richer'' property leads to a flat prior. As we show later, our formulation allows to retrieve such flat priors and thus generalizes the Uniform Process.

\subsection{Contributions}
In the present work, we derive the Powered Chinese Restaurant Process (PCRP) that allows controlling the ``rich-get-richer'' property while generalizing state-of-the-art works --- rich-get-no-richer (Uniform process), rich-get-less-richer, ``rich-get-richer'' (DP), and rich-get-more-richer. Doing so, we define the Powered Dirichlet-Multinomial distribution. We detail some key-properties of the Powered Dirichlet Process (convergence, expected number of clusters). Finally, we show that controlling the ``rich-get-richer'' prior of simple models yields better results on synthetic and real-world datasets.



\section{The model}
\subsection{The Dirichlet-Multinomial distribution}
We recall:
\begin{equation}
\label{eq-Mult}
    Dir(\vec{p} \vert \vec{\alpha}) = \frac{\prod_k p_k^{\alpha_k - 1}}{B(\vec{\alpha})} \ \ \ \, \ \ \ \ Mult(\vec{N} \vert N, \vec{p}) = \frac{\Gamma(\sum_k N_k + 1)}{\prod_k \Gamma(N_k + 1)} \prod_k p_k^{N_k} 
\end{equation}
With $\vec{N} = (N_1, N_2, ..., N_K)$ where $N_k$ is the integer number of draws assigned to cluster $k$, $N = \sum_k N_k$ the total number of draws, $\Gamma(x)=(x-1)!$ and $B(\vec{x}) = \prod_k \Gamma(x_k)/\Gamma(\sum_k x_k)$.

The regular Dirichlet process can be derived from the Dirichlet-Multinomial distribution. The Dirichlet-Multinomial distribution is defined as follows:
\begin{equation}
\label{eq-DirMult}
\begin{split}
    p(\vec{N} \vert \vec{\alpha}, n) &= \int_{\vec{p}} p(\vec{N}\vert \vec{p}, n)p(\vec{p} \vert \vec{\alpha}) d\vec{p}\\
    &= \frac{(n!)\Gamma(\sum_k \alpha_k)}{\Gamma(n+\sum_k \alpha_k)} \prod_{k=1}^{K}\frac{\Gamma(N_k + \alpha_k)}{(N_k!)\Gamma(\alpha_k)}\\
    \text{where }&\Vec{p} \sim Dir(\vec{p} \vert \vec{\alpha}) \, ; \, \Vec{N} \sim Mult(\vec{N} \vert n, \Vec{p})
\end{split}
\end{equation}
In Eq.\ref{eq-DirMult}, we sample $n$ values over a space of $K$ distinct clusters each with probability $\vec{p}=(p_1, p_2, ..., p_K)$, using a Dirichlet prior with parameter $\vec{\alpha} = (\alpha_1, \alpha_2, ..., \alpha_K)$. As we will show in the next section, we can derive the Dirichlet process equation by iterating the Dirichlet-Multinomial distribution. More precisely, one has to compute a new observation's conditional distribution to belong to any cluster given the allocation of all the previous random variables when $K \rightarrow \infty$.

\subsection{Powered conditional Dirichlet prior}
In the derivation of the standard Dirichlet-Multinomial posterior predictive, one considers a categorical distribution coupled with a Dirichlet prior on its parameter $\vec{p}$. Usually, this prior is linearly dependent on previous draws from the distribution. We propose to modify this assumption by using a Dirichlet prior that depends on the history of draws as:
\begin{equation}
\label{eq-DirMultPrior}
    Dir_r(\vec{p} \vert \vec{\alpha}, \vec{N}) = \frac{1}{B(\vec{\alpha} + \vec{N}^r)} \prod_k p_k^{\alpha_k + N_k^r - 1}
\end{equation}
In Eq.\ref{eq-DirMultPrior}, the vector $\vec{N}^r$ shifts the parameter $\vec{\alpha}$ according to the count of draws allocated to each cluster $k$ up to the n$^{th}$ draw. The parameter $r \in \mathbb{R}^+$ controls the intensity of this shift for each entry of $\vec{X}$.

We demonstrate that the Powered Dirichlet distribution is a conjugate prior of the Multinomial distribution, by writing Eq.\ref{eq-DirMultPrior} as:
\begin{equation}
\begin{split}
\label{eq-DirMultPrior2}
    Dir_r(\vec{p} \vert \vec{\alpha}, \vec{N}) &= \frac{1}{B(\vec{\alpha} + \vec{N}^r)} \prod_k p_k^{\alpha_k - 1} \prod_k p_k^{N_k^r}\\
    &\stackrel{\text{Eqs.\ref{eq-Mult}}}{=} \frac{B(\vec{\alpha}) \prod_k N_k^r!}{B(\vec{\alpha} + \vec{N}^r) (\sum_k N_k^r)!} Dir(\vec{p} \vert \vec{\alpha}) Mult(\vec{N}^r \vert \sum_k N_k^r, \vec{p})\\
\end{split}
\end{equation}
where the prior on vector $\vec{N}$ is a regular Multinomial distribution of parameter $N=\sum_k N_k^r$. Note that for certain values of $r$, the vector $\vec{N}^r$ might not be made of integer values; the resulting Multinomial prior on $\vec{N}^r$ must then be expressed in terms of $\Gamma$ functions (see Eq.\ref{eq-Mult}) to be valid for $\vec{N}^r \in \mathbb{R}^{\vert \vec{N} \vert}$. 
Distributions of non-integer counts are not new in the literature \cite{Ghitza2013,McCarthy2012DifferentialEA,Khurshid2005ConfidenceIntervNegBinDist} and are essentially allowed by the generalized definition of the factorial function in terms of the gamma function. 
When $r=1$, we recover the standard Dirichlet-Multinomial prior on $\vec{p}$ for the $n^{th}$ draw; the history of draws $\vec{N}$ can be expressed as the result of $N$ independent draws of equal probability $\vec{p}$. When $r \neq 1$, the prior on $\vec{N}$ is sampled from a Multinomial distribution in which the number of samples drawn depends on $r$ as $\sum_k N_k^r$. For instance, let $\vec{N} = (1, 2)$ and $r=2$: the resulting powered conditional Dirichlet prior would then be sampled from a Multinomial distribution $Mult(\vec{N}=(1,4) \vert N=5, \vec{p}=(p, 1-p))$.

\subsection{Posterior predictive}
We now derive the posterior distribution for the $n^{th}$ draw to belong to a cluster $c$ given all previous draws. We assume that $\vec{C_-}$ represents all previous realizations up to $n-1$, that is, the cluster to which each previous draw has been associated. For simplicity of notation, we define the population of a cluster $k$ at time $n-1$ as $N_k = \vert \{C_i \vert i=k\}_{i=1, 2, ..., n-1} \vert$. We are now looking at the probability distribution of its $n^{th}$ draw to belong to $c$. It is expressed as the probability of a draw from the categorical distribution given all previous observations (because there is only one new draw, it is the same as a Multinomial distribution with parameter $N=1$) combined with the powered Dirichlet prior defined Eq.\ref{eq-DirMultPrior}. Then:
\begin{equation}
    \label{eq-DirCatr}
    \begin{split}
        DirCat_r(C_n = c \vert \vec{\alpha}, \vec{C_-}) &= \int_{\vec{p}} Cat(C_n=c \vert \vec{p}) \underbrace{Dir_r(\vec{p} \vert \vec{\alpha}, \vec{N})}_{\textbf{Eq.\ref{eq-DirMultPrior}}} \\
        =& \int_{\vec{p}} \frac{1}{B(\vec{\alpha} + \vec{N}^r)} \prod_k p_k^{c_k + \alpha_k + N_k^r - 1}\\
        &= \frac{B(\vec{c} + \vec{\alpha} + \vec{N}^r)}{B(\vec{\alpha} + \vec{N}^r)}
    \end{split}
\end{equation}
where $\vec{c}$ is a vector of the same length as $\vec{\alpha}$ and $\vec{C}$ whose c$^{th}$ entry equals 1, and 0 anywhere else. Alternative demonstrations of this result are possible \cite{Sethuraman1994,Wilks1992}.

\subsection{Powered Chinese Restaurant process}
We finally derive an expression for the Powered Chinese Restaurant process from Eq.\ref{eq-DirCatr}. We recall that $N_k = \vert \{C_{-i} \vert i=k\}_{i=1, 2, ..., n-1} \vert$. Taking back the conditional probability for the $n^{th}$ observation to belong to cluster $c$ (Eq.\ref{eq-DirCatr}), we have:
\begin{equation}
\label{eq-derivPCRP}
\begin{split}
    p(C_n = c \vert \vec{C_-}, \vec{\alpha}) =& DirCat_r(C_n=c \vert \vec{C_-}, \vec{\alpha})\\
    =& B(\vec{c}+\vec{N^r}+\vec{\alpha}) / B(\vec{N^r}+\vec{\alpha})\\
    =& \Gamma(N_c^r + \alpha_c + 1) \frac{\prod_{k \neq c} \Gamma(N_k^r + \alpha_k)}{\Gamma(1 + \sum_k N_k^r + \alpha_k)} \frac{\Gamma(\sum_k N_k^r + \alpha_k)}{\prod_{k \neq c} \Gamma(N_k^r + \alpha_k)}\\
    =& \frac{(N_c^r + \alpha_c)}{\sum_k N_k^r + \alpha_k} \frac{\prod_{k} \Gamma(N_k^r + \alpha_k)}{\Gamma(\sum_k N_k^r + \alpha_k)} \frac{\Gamma(\sum_k N_k^r + \alpha_k)}{\prod_{k} \Gamma(N_k^r + \alpha_k)}\\
    =& \frac{N_c^r + \alpha_c}{\sum_k N_k^r + \alpha_k}
\end{split}
\end{equation}

Finally, taking the limit $K \rightarrow \infty$ and defining $\sum_k \alpha_k \stackrel{\rightarrow}{K \rightarrow \infty} \alpha$, we find the Powered Chinese Restaurant Process:
\begin{equation}
\label{eq-PowCRP}
    P-CRP (C_i = c \vert C_1, C_2, ..., C_{i-1}) = 
    \begin{cases}
    \frac{N_c^r}{\alpha + \sum_k^K N_k^r} \text{ if c = 1, 2, ..., K}\\
    \frac{\alpha}{\alpha + \sum_k^K N_k^r} \text{ if c = K+1}
    \end{cases}
\end{equation}

The formal derivation of the Powered Chinese Restaurant process in Eq.\ref{eq-PowCRP} and the demonstration of its link to the conditional Dirichlet prior on $\vec{p}$ are the first main contribution of this work. Besides, this demonstration uncovers the link between the prior in Eq.\ref{eq-DirMultPrior} and an exotic formulation of the Multinomial distribution, which has never been considered before.
As stated in the introduction, special cases of the process have already been used in some occasions \cite{wallach2010,Qin2003,Jensen2008} but never demonstrated. Furthermore, this formulation generalizes the Uniform process when $r\rightarrow 0$ \cite{wallach2010}, the Dirichlet process when $r \rightarrow 1$ and the Pitman-Yor process when $r_k(N_k) = (log(1-\beta/N_k) + log(N_k))/log(N_k)$ and $\alpha(K) = \alpha + \beta K$ (see Eq.\ref{eq-PY}, we recall that $e^{\log x} = x$). The present expression explicitly allows for controlling the importance of the ``rich-get-richer'' property as well as recovering state-of-the-art processes. 

\begin{figure}
    \centering
    \includegraphics[width = 0.8\columnwidth]{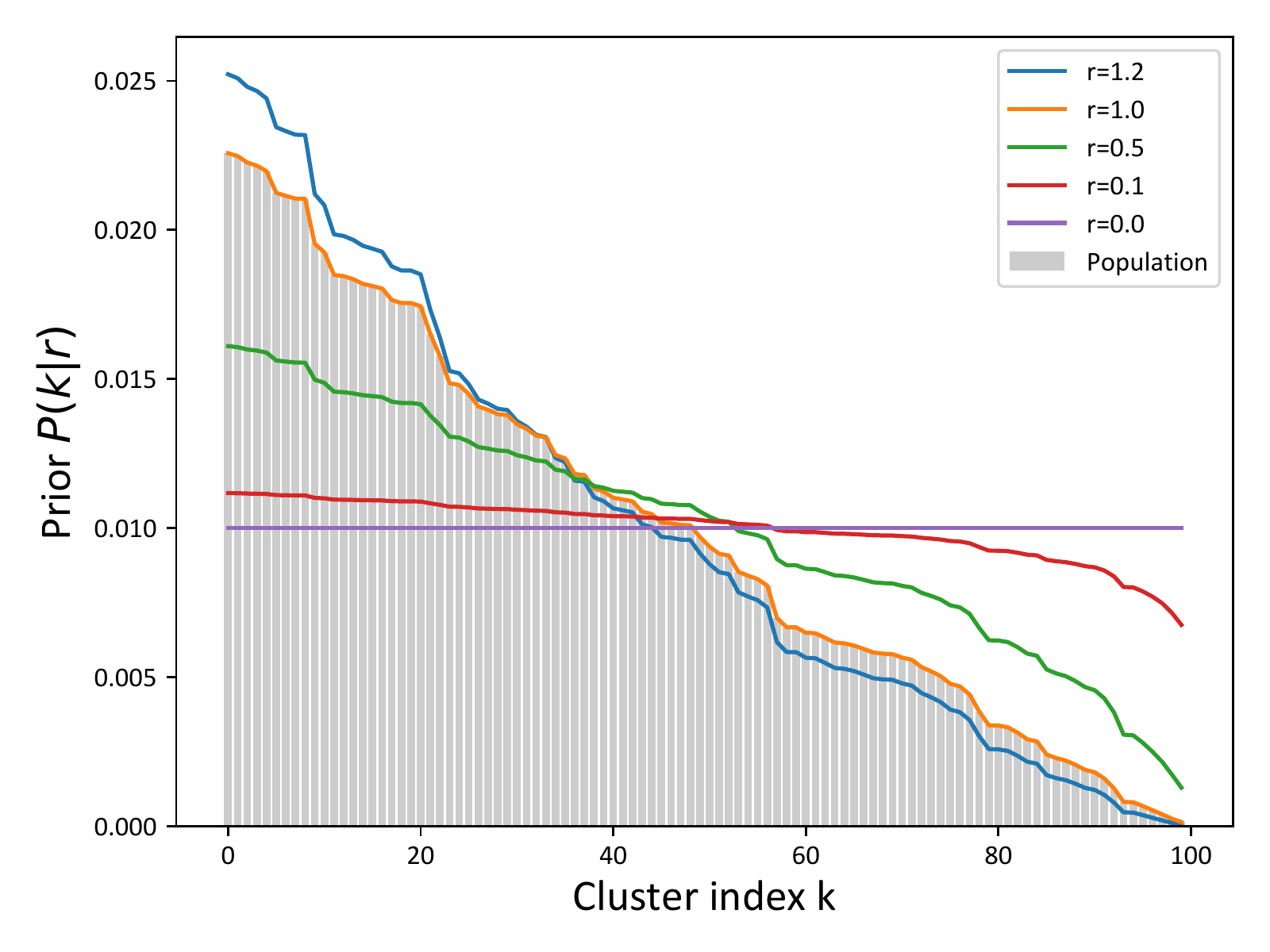}
    \caption{Effect of $r$ on the Powered Chinese Restaurant process prior probability.}
    \label{fig-prior}
\end{figure}

We illustrate the change on prior probability for an existing cluster to get chosen induced by the Powered Chinese Restaurant process in Fig.\ref{fig-prior} -- we do not plot the prior probability for a new cluster to be created. This figure plots the population of clusters (grey bars) and their associated prior probability of getting chosen. When $r>1$, the most populated clusters are associated with a more significant prior probability than in the standard CRP, whereas the less populated ones have even less chances to get chosen; rich-get-more-richer, the prior on population is more peaky on large clusters. On the other hand, when $r<1$, most populated clusters have less chances to get chosen than in CRP, whereas less populated ones have an increased chance of getting chosen; rich-get-less-richer, the prior on population is flatter across clusters of different sizes. In the limit case $r=0$, the clusters' population does not play any role anymore; rich get-no-richer, the prior is flat over all clusters. Note that if we wanted to represent the Pitman-Yor process prior in this figure, it would correspond to the plot for $r=1$ vertically shifted of $-\beta$ (such as defined Eq.\ref{eq-PY}) leading to an increased probability of creating a new cluster of $\beta K$ (not represented in the plot) \cite{PitmanYor1997}. Varying the parameter $\alpha$ of $\Delta \alpha$ plays a similar role as $\beta$ in this situation. It would uniformly shift the prior probability for each existing cluster to get chosen by $\frac{\Delta \alpha}{K}$ and increase the probability of creating a new one by $\Delta \alpha$. For Both Pitman-Yor and Dirichlet processes, the linear dependence of each cluster's population does not change.

In Fig\ref{fig-prior}, we understand that the Powered Chinese Restaurant process allows for defining priors from clusters population that are not possible when tuning the Chinese Restaurant or Pitman-Yor processes. Introducing non-linearity in the dependence on previous observations allows giving any importance to the ``rich-get-richer'' property.

\section{Properties of the Powered Chinese Restaurant process}
We will now investigate some key properties of the Powered Chinese Restaurant process. We recall that $N_k$ is the population of the cluster $k$, and $N = \sum_k N_k$.
\subsection{Convergence}
\begin{prop}
\label{th-statio}
For $N \rightarrow \infty$, the Powered Chinese Restaurant process converges towards a stationary distribution. When $r<1$, it converges towards a uniform distribution over all the possible clusters, and when $r>1$, it converges towards a Dirac distribution on a single cluster.
\end{prop}
\begin{proof}
We consider a simple situation where only 2 clusters are involved. The generalization to the case where $K$ clusters are involved is straightforward. When clusters' population is large enough, we make the following Taylor approximation:
\begin{equation}
\label{eq-taylorapprox}
    \begin{split}
        (N_i+1)^r &= N_i^r(1 + \frac{1}{N_i^r}) = N_i^r + r N_i^{r-1} + \mathcal{O}(N^{r-2})
    \end{split}
\end{equation}

Since the population of a cluster $N_i$ is a non-decreasing function of $N$, we assume that first order Taylor approximation holds when $N \rightarrow \infty$. Given clusters population at the $N^{th}$ observation, we perform a stability analysis of the gap between probabilities $\Delta p(N) = p_1(N)-p_2(N)$. We recall that the probability for cluster $i$ to get chosen is $p_i(N) = N_i^r / (\sum_k N_k^r)$ and that either of the clusters is chosen with this probability at the next step (at step $N+1$, $\Delta p(N+1) = p_1(N+1)-p_2(N)$ with probability $p_1(N)$ and $\Delta p(N+1) = p_1(N)-p_2(N+1)$ with probability $p_2(N)$). Explicitly the variation of the gap between probabilities when $N$ grows is written as:
\begin{equation}
\label{eq-varGap}
    \begin{split}
        &\frac{p_1(N) (p_1(N+1) - p_2(N)) +  p_2(N) (p_1(N) - p_2(N+1)) - \Delta p(N)}{\Delta p(N)}\\
        \stackrel{\text{Eq.\ref{eq-taylorapprox}}}{\approx} &\frac{1}{p_1(N) - p_2(N)} \times \left( p_1(N) \frac{N_1^r - N_2^r + r N_1^{r-1}}{N_1^r + N_2^r + r N_1^{r-1}}  + p_2(N) \frac{N_1^r - N_2^r - r N_2^{r-1}}{N_1^r + N_2^r + r N_2^{r-1}} \right)\\
        = &\frac{2 r N_1^r N_2^r}{(N_1^r+N_2^r+rN_1^{r-1})(N_1^r+N_2^r+rN_2^{r-1})}\left( \frac{N_1^{r-1} - N_2^{r-1}}{N_1^r-N_2^r} \right)
    \end{split}
\end{equation}

We see in Eq.\ref{eq-varGap} that the sign of the variation of the gap between probabilities depend only on the term $\frac{N_1^{r-1} - N_2^{r-1}}{N_1^r-N_2^r}$. We can therefore perform a stability analysis of the Powered Chinese Restaurant process using only this expression.

When $0<r<1$, the following relation holds: $N_1^{r-1} - N_2^{r-1} < 0 \Leftrightarrow N_1^{r} - N_2^{r} > 0 \ \forall N_1, N_2$; that makes right hand side of Eq.\ref{eq-varGap} negative. Therefore adding a new observation statistically reduces the gap between the probabilities of the two clusters. We could forecast this prediction from Eq.\ref{eq-taylorapprox} by seeing that adding a new observation to a large cluster increases its probability to get chosen lesser than for a small cluster -- rich-get-less-richer. Moreover, we see from Eq.\ref{eq-taylorapprox} that a crowded cluster (such as $N_1^r \gg N_2^r$) see its probability evolve as $N^{r-1}$. Asymptotically, the only fixed point of Eq.\ref{eq-varGap} when $N \rightarrow \infty$ is $N_1 \rightarrow N_2$, which implies a uniform distribution. 

On the contrary, when $r>1$ we have the following relation: $N_1^{r-1} - N_2^{r-1} > 0 \Leftrightarrow N_1^{r} - N_2^{r} > 0 \ \forall N_1, N_2$; ; that makes right hand side of Eq.\ref{eq-varGap} positive. Adding a new observation statistically increases the gap between probabilities. From Eq.\ref{eq-taylorapprox}, we see that adding an observation to a large cluster increases its probability with its population -- rich-get-more-richer. In this case, Eq.\ref{eq-varGap} has $K+1$ fixed points, with $K$ the number of clusters. The uniform distribution is an unstable fixed point, while $K$ Dirac distributions (each on one cluster) are stable fixed points of the system. It means the gap converges to $1$, that is a probability of 1 for one cluster and a probability of 0 for the others. 

When $r=1$, the right hand side of Eq.\ref{eq-varGap} is null. It means the gap remains statistically constant $\forall N_i$, which is a classical result for the regular Dirichlet process. This convergence has already been studied on many occasions \cite{Ferguson1973,Arratia1992}. 

We note that as $r \rightarrow 0$, Eq.\ref{eq-varGap} is not defined anymore. That is because the probability for a cluster to be chosen does not depend on its population anymore. In this case, $p_1(N) - p_2(N) \propto N_1^0 - N_2^0 = 0$: the probability for any cluster to be chosen is equal, hence the Uniform process -- ``rich-get-no-richer''.

\qed
\end{proof}

\subsection{Expected number of tables}

\begin{prop}
\label{th-slowVarp}
When $N$ is large, $\sum_k N_k^r$ varies with $N$ as $N^{\frac{r^2+1}{2}}$ when $r<1$, and with $N^{r}$ when $r\geq 1$.
\end{prop}
\begin{proof}
Taking back Eq.\ref{eq-PowCRP}, we are interested in the variation of $p_i = \frac{N_i^r}{\sum_k N_k^r}$ according to $N$ when $N_i^r$ is large:
\begin{equation}
\label{eq-slowVarp}
\begin{split}
p_i(N+1)-p_i(N) \approx \begin{cases}
\frac{rN_i^{r-1} + \mathcal{O}(N^{r-2})}{\sum_k N_k^r} &\text{ if $N_i$ grows}\\
0 &\text{ else}
\end{cases}
\end{split}
\end{equation}

We see in Eq.\ref{eq-slowVarp} that for $r<1$, the larger $N_i$ the slower the variation of $p_i$. It means that for large $N_i^r$, we can write $N_i \propto N p_i$, with $p_i$ a constant of $N$. Since $N_i$ is either way a non-decreasing function of $N$, we reformulate the constraint $N_i^r$ large in $N^r$ large.

For $r>1$, the probability $p_i$ varies greatly with $N$ and quickly converges to 1 for large $N$ (see Proposition \ref{th-statio}), and so $N_i \approx N$ for cluster $i$ and $N_{j \neq i} \ll N_i \ \forall j$.

Since the sum $\sum_k N_k^r$ essentially varies according to large $N_k$, we can approximate $\sum_k N_k^r \approx N^r \sum_k p_k^r$ for large $N^r$.

Besides, we showed in Proposition \ref{th-statio} that for large $N$ the process converges towards a uniform distribution for $r<1$ and towards a Dirac distribution when $r>1$. Therefore, we can express $\sum_k^K p_k^r$ as:
\begin{equation}
    \sum_k^K p_k^r \stackrel{N \gg 1}{\approx}
    \begin{cases}
    K^{1-r} &\text{for $r<1$}\\
    1 &\text{for $r \geq 1$}
    \end{cases}
\end{equation}

Based on the demonstration of Eq.4 in \cite{wallach2010}, we suppose that $K$ evolves with $N$ as $N^{\frac{1-r}{2}}$ when $r<1$. We verify that this assumption holds in the Experiment section.

Therefore, we can write:
\begin{equation}
    \sum_k N_k^r \approx N^r \sum_k^K p_k^r \approx \begin{cases}
    N^{r} \left( N^{\frac{1-r}{2}} \right)^{1-r} = N^{\frac{1+r^2}{2}} &\text{for $r<1$}\\
    N^{r} &\text{for $r \geq 1$}\\
    \end{cases}
\end{equation}

\qed
\end{proof}

\begin{prop}
\label{th-expK}
The expected number of tables of the Powered Chinese Restaurant process evolves with $N \gg 1$ as $H_{\frac{r^2+1}{2}}(N)$ for $r<1$ and as $H_{r}(N)$ when $r \geq 1$, where $H_m(n)$ is the generalized harmonic number.
\end{prop}
\begin{proof}
In general, the expected number of clusters at the $N^{th}$ step can be written as:
\begin{equation}
\label{eq-expK}
    \mathbb{E}(K \vert N, r) = \sum_1^N \frac{\alpha}{\sum_k N_k^r + \alpha} \stackrel{N^r \gg 1}{\propto} \sum_1^N \frac{1}{\sum_k N_k^r}
\end{equation}

We showed in Proposition \ref{th-slowVarp} that we can rewrite $\sum_k N_k^r \propto N^{\frac{r^2 + 1}{2}}$ when $r<1$ and $\sum_k N_k^r \propto N^r$ when $r\geq 1$.
Injecting this result in Eq.\ref{eq-expK} for $r$, we get:
\begin{equation}
    \label{eq-expK-fin}
    \mathbb{E}(K \vert N, r) \stackrel{N^r \gg 1}{\propto}
    \begin{cases} \sum_1^N\frac{1}{N^{\frac{r^2 + 1}{2}}} = H_{\frac{r^2+1}{2}}(N)\\
    \sum_1^N\frac{1}{N^r} = H_{r}(N)
    \end{cases}
\end{equation}

\qed
\end{proof}

For $r=1$, $\mathbb{E}(K \vert N, r=1) \propto H_1(N) \approx \gamma + \log(N)$ where $\gamma$ is the Euler–Mascheroni constant, which is a classical result for the regular Dirichlet process. 

When $r>1$ and $N \rightarrow \infty$, the term $H_{\frac{r^2+1}{2}}(N)$ converges towards a finite value and the sum $\sum_k p_k^r$ goes to 1 (see Proposition \ref{th-statio}). By definition $\mathbb{E}(K \vert N, r>1) \stackrel{N \rightarrow \infty}{\propto} \zeta(\frac{r^2+1}{2})$, where $\zeta$ is the Riemann Zeta function.

When $r < 1$, we can approximate the harmonic number in a continuous setting. We rewrite Eq.\ref{eq-expK-fin} as:
\begin{equation}
    \begin{split}
    \mathbb{E}(K \vert N, r) &\stackrel{N^r \gg 1}{\propto} \sum_{n=1}^N\frac{1}{n^{\frac{r^2 + 1}{2}}}
    \stackrel{N^r \gg 1}{\approx} \int_1^N n^{-\frac{r^2 + 1}{2}} dn
    = \frac{2}{1 - r^2}(N^{\frac{1-r^2}{2}}-1) 
    \end{split}
\end{equation}
One can show that $\frac{N^{1-x}-1}{1-x} = H_x(N) + \mathcal{O}(\frac{1}{N^x})$. Therefore, the Powered Chinese Restaurant process exhibits a power-law behaviour similar to the Pitman-Yor process Eq.\ref{eq-PY} for $r = \sqrt{1 - 2\beta}$ for $0<r<1$. For values of $r>1 \Leftrightarrow \beta<0$, the equivalent Pitman-Yor process is not defined unlike the Powered Chinese Restaurant process. Note that there is \textit{a priori} no reason for $r$ to be constrained in the domain of real number. Complex analysis of the process might be an interesting lead for future works.

\section{Experiments}
\label{Experiments}
\subsection{Numerical validation of propositions}
\begin{figure}
    \centering
    \includegraphics[width = \textwidth]{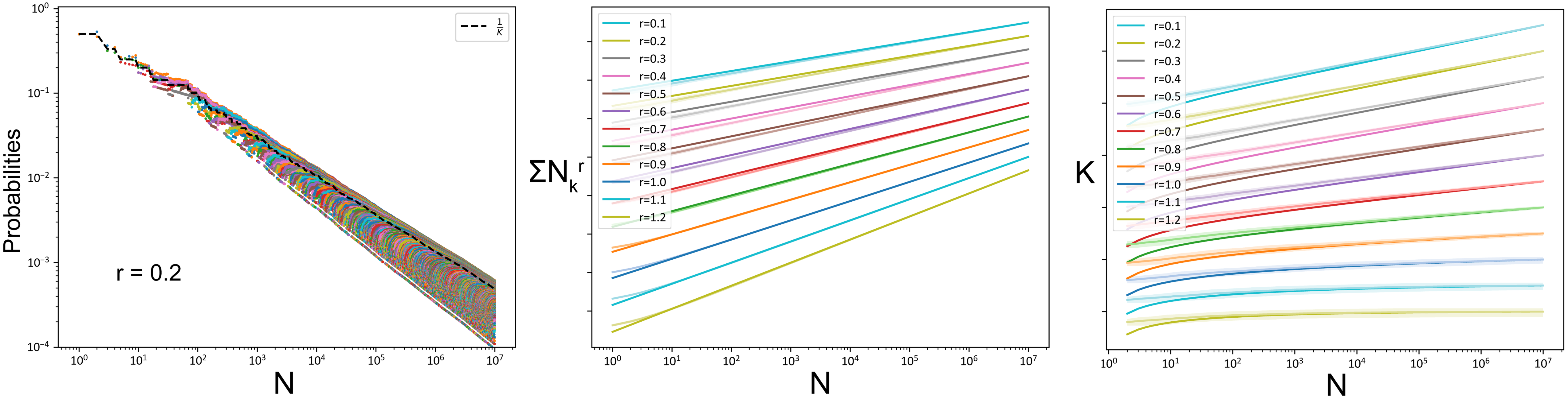}
    \caption{Numerical validation of Propositions \ref{th-statio} (left), \ref{th-slowVarp} (middle), \ref{th-expK} (right). In the third and fourth plots, the theoretical results are the solid lines and the associated numerical results are the transparent lines of same color. Except for small $N$, the difference between theory and experiments is almost indistinguishable.}
    \label{fig-validTh}
\end{figure}

First of all, we present numerical confirmations of propositions stated above (Propositions \ref{th-statio}, \ref{th-slowVarp}, \ref{th-expK}) by simulating 100 independent Powered Chinese Restaurant processes with parameter $\alpha=1$ for various values of $r$. We present the results of numerical simulations in Fig.\ref{fig-validTh}.

On the \textbf{left} part, we plot the evolution of the probability for each cluster to be chosen as $N$ grows for $r=0.2$ for one run. We see that the probabilities do not remain constant but instead diminish as the number of clusters grows. The figure suggests they all converge to a common value (a uniform probability) as shown in Proposition \ref{th-statio}. The black line shows the probability of a uniform distribution. We chose not to show the results for $r>1$; in this case, one probability goes to 1 as the other fades to 0 as $N$ grows, as expected.

In the \textbf{middle} part of the figure, we plot the expression for $\sum_k N_k^r$ derived in Proposition \ref{th-slowVarp} (solid lines) versus the value of the sum from experimental results (transparent lines), averaged over 100 runs. Note that plots are in a log-log scale and that curves have been shifted vertically for visualization purposes. As assumed in Proposition \ref{th-slowVarp}, the approximation holds for all values of $r$.

Finally in the \textbf{right} picture, we plot the evolution of the number of clusters $K$ versus $N$ according to Proposition \ref{th-expK} (solid lines) and experiments (transparent lines). The error bars correspond to the standard deviation over the 100 runs. We see that the expression derived in Proposition \ref{th-expK} accounts well for the evolution of the number of clusters. Note that plots are in a log-log scale and that curves have been shifted vertically for visualization purposes. We must point out that there is a constant shift from experiments to the theory that does not appear on the plot (because of the rescaling). This shift comes from the approximation of large $N^r$ which is not valid at the beginning of the process. However, it does not play any role in the evolution of $K$ as $N$ grows large enough.

\subsection{Use case: infinite Gaussian mixture model}
We now illustrate the usefulness of a prior that alleviates the ``rich-get-richer'' property with specific synthetic datasets and with a real-world application. We choose to consider as an illustration its use as a prior in the infinite Gaussian mixture model\footnote{All codes and datasets can be found at https://anonymous.4open.science/r/91ea587e-fba6-4ba0-887e-79d87abf0b31/}. We choose this application to ease visual understanding of the implications of the P-CRP, but the argument holds for other models using DP priors as well (text modeling, gene expression clustering, etc.). 

We consider a classical infinite Gaussian mixture model coupled with a Powered Dirichlet process prior. We fit the data using a standard collapsed Gibbs sampling algorithm for IGMM \cite{Rasmussen1999,wallach2010,Yin2014}, with a Normal Inverse Wishart prior on the Gaussians' parameters. The input data is shuffled at each iteration to reduce the ordering bias from the dataset. Note that we cannot completely get rid of the bias because the Powered Dirichlet Process is not exchangeable for all $r$. The problem has been addressed on numerous occasions (Uniform process \cite{wallach2010}, distance-dependent CRP \cite{Blei2011,Ghosh2014}, spectral CRP \cite{Socher2011}) and shown to induce negligible variations of results in the case of Gibbs sampling. We stop the sampler once the likelihood of the model reaches stability
; we repeat this procedure 100 times for each value of $r$. 
Finally, the parameter $\alpha$ is set to 1 in all experiments (see Section \ref{motivation}). 

\subsubsection{Synthetic data}
\begin{figure}
    \centering
    \includegraphics[width = \columnwidth]{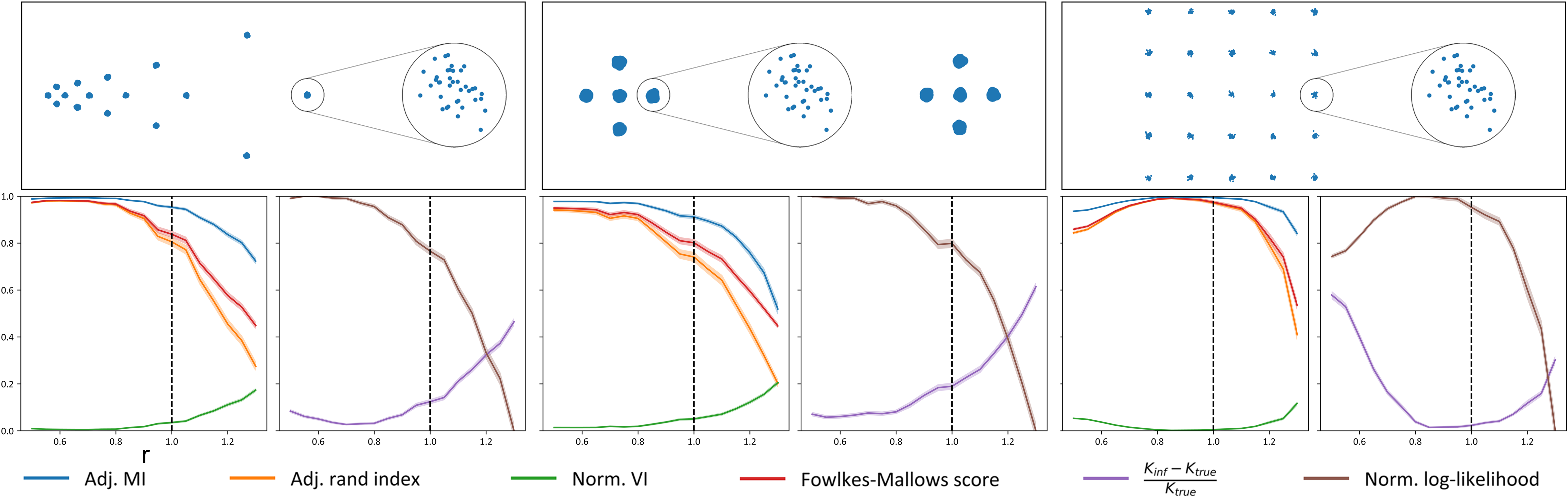}
    \caption{Application for synthetic data. (Top) Original datasets used for the experiments. (Bottom) Results for various values of $r$; the x and y axes all the same. The dashed line indicates the regular DP prior as $r=1$. The error correspond to the standard error of the mean over all runs.}
    \label{fig-Results}
\end{figure}
We present the results on synthetic data in Fig.\ref{fig-Results}. We consider standard metrics in clustering evaluation with a non-fixed number of clusters: mutual information score and rand index both adjusted for chance, normalized variation of information, Fowlkes-Mallow score, marginal likelihood (normalized for visualization) and absolute relative variation of the inferred number of clusters according to the number used in the generation process. Note that we purposely chose stereotypical cases to illustrate the argument better. The dataset on the \textbf{left} of Fig.\ref{fig-Results} is informative about the change induced by $r$. Here, clusters are distributed at various scales in the dataset; we see that the lower the value of $r$, the better the results. Indeed, when $r$ is small, the model can distinguish clusters in the dense area better, whereas when $r$ is closer to 1, the clusters in the dense area are put together in a larger cluster. The same happens with the dataset in the \textbf{middle} of Fig.\ref{fig-Results}, where clusters are distributed according to two different scales. Finally, on the \textbf{right} part of Fig.\ref{fig-Results}, we see an optimum $r$ exists to distinguish the clusters distributed on a grid; it makes sense since only one scale in clusters distribution is involved in this dataset.

\subsubsection{Real data}
\begin{figure}
    \centering
    \includegraphics[width = \columnwidth]{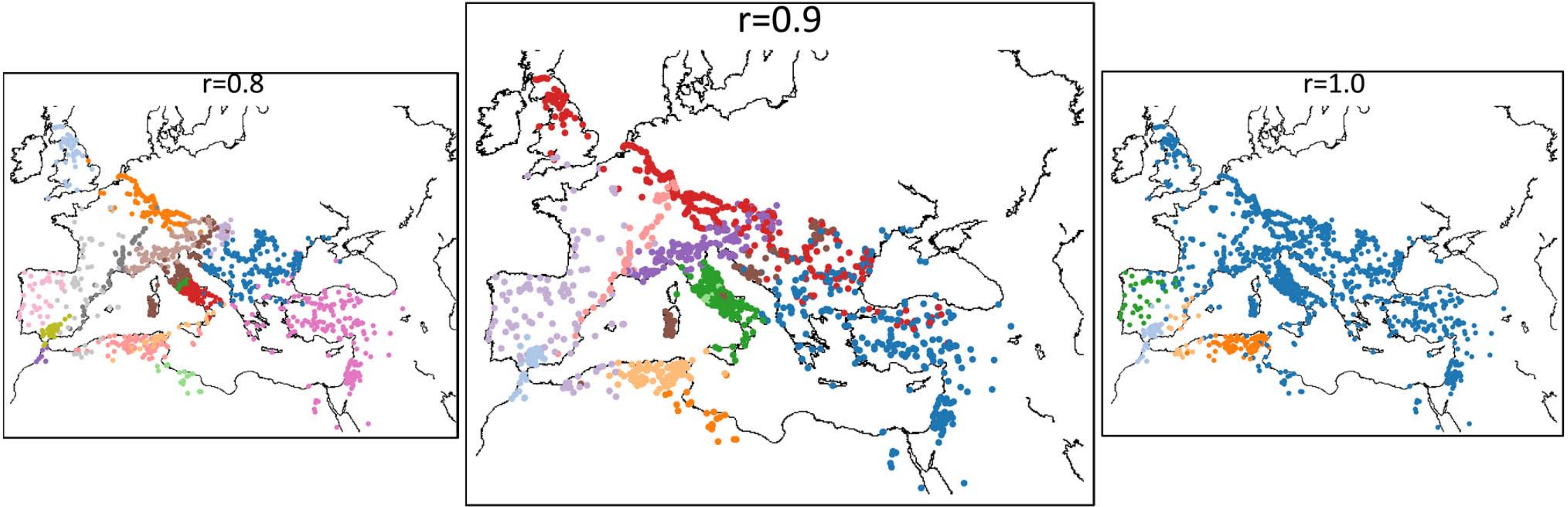}
    \caption{Application to spatial clustering on geolocated data for $r=0.8$ (left), $r=1$ (right) and $r=0.9$ (middle). We see that the Powered Chinese Restaurant process for $r=0.9$ and $r=0.8$ describes the data better than the classical one for $r=1$.}
    \label{fig-Antoni}
\end{figure}
We now illustrate the interest of using an alternate form of prior for the Infinite Gaussian Mixture model on real-world data. We consider a dataset of 4.300 roman sepulchral inscriptions comprising the substring ``Antoni'' that have been dated between 150AC and 200AC and assigned with map coordinates. The dates correspond to the reign of Antoninus Pius over the Roman empire. The dataset is available on Clauss-Slaby repository\footnote{http://www.manfredclauss.de/fr/index.html}. It was common to give children or slaves the name of the emperor; the dataset gives a global idea of the main areas of the roman empire at that time \cite{Hanson2017}. The task here is to discover spatial clusters of individuals named after the emperor. We expect to find geographical clusters around: Italy, Egypt, Gauls, Judea, and all along the \textit{limes} (borders of the roman empire, which concentrate lots of sepulchral inscriptions for war-related reasons) \cite{Hanson2016}. We present the results for various values of $r$ in Fig.\ref{fig-Antoni}.

We see that when $r=1$, the classical CRP prior is not fit for describing this dataset, as it misses most of the clusters. On the other hand, when $r=0.9$, the infinite Gaussian mixture model retrieves the expected clusters. It also makes some clusters that were not expected, such as the north Italian cluster or the long cluster going through Spain and France that corresponds to roman roads layout (via Augusta and via Agrippa; it was common to bury the dead on roads edges). Finally, when $r=0.8$, we get even more detail: some of the main clusters are broken into smaller ones (Italy breaks into Rome, North Italy, and South Italy; Britain becomes an independent cluster, etc.). In this case, changing $r$ controls the level of details of the clustering. We see how different results can be according to the extent the model relies on the ``rich-get-richer'' prior and how it is needed to control it to make modeling relevant to every situation.

\section{Conclusion}
In this article, we discuss the necessity of controlling the ``rich-get-richer'' property that arises from the common Chinese Restaurant Process usual formulation. We discuss cases where this modeling hypothesis must be alleviated or strengthened to describe data more accurately. To this end, we derive the Powered Chinese Restaurant Process from a powered version of the Dirichlet-Multinomial distribution. This formulation allows reducing the expected number of clusters, which is not possible in the standard Pitman-Yor processes, while generalizing the standard Dirichlet process and the Uniform process. The principal feature of this formulation is that it allows for direct control of the ``rich-get-richer'' priors' importance. We derive elementary results on convergence and the expected number of clusters of the new process. Finally, we show that it yields better results on synthetic data and illustrates a possible use case with real-world data. For future works, it might be interesting to investigate cases where $r$ takes non-positive values (which might lead to a ``poor-get-richer'' kind of process) or complex values (for $r=a+ib$ the prior probability $N^r$ would have an amplitude $N^a$ and a phase $b \log N$).

The regular Chinese Restaurant process has been used for decades as a powerful prior in many real-world applications. However, alternate forms for this prior have been little explored. It would be interesting to consider the changes brought to state-of-the-art models by varying the importance of the ``rich-get-richer'' prior as proposed in this paper.

%
%
%
\bibliographystyle{splncs04}
\bibliography{mybibliography}

\end{document}


\title{Supplementary Material for Powered Dirichlet Process for Modulating Dirichlet Priors Informativeness}
%
%
%
%
%
\maketitle              

\section{Powered Dirichlet-Multinomial distribution}
As a side result, we derive a closed-form expression for the Powered Dirichlet-Multinomial distribution based on our definition of the powered conditional Dirichlet prior:
\begin{equation}
    \begin{split}
        \label{eq-DirMultr}
        DirMult_r(\vec{N} \vert \vec{\alpha}) &= \int_{\vec{p}} Mult(\vec{N}^r \vert \vec{p}) Dir(\vec{p} \vert \vec{\alpha}) d\vec{p}\\
        &= \int_{\vec{p}} \frac{ (\sum_k N_k^r)!}{\prod_k N_k^r! B(\vec{\alpha})} \prod_k p_k^{\alpha_k + N_k^r - 1} d\vec{p}\\
        &=  \frac{ (\sum_k N_k^r)!}{\prod_k N_k^r! B(\vec{\alpha})} B(\vec{\alpha} + \vec{N}^r) \underbrace{\int_{\vec{p}}  Dir(\vec{p} \vert \alpha_k + N_k^r) d\vec{p}}_{=1} \\
        &= \frac{B(\vec{\alpha} + \vec{N}^r) (\sum_k N_k^r)!}{B(\vec{\alpha}) \prod_k N_k^r! } 
    \end{split}
\end{equation}

\bibliographystyle{splncs04}
\bibliography{mybibliography}